# On the fractal nature of mutual relevance sequences in the Internet news message flows

*S. Braichevsky[1], D. Lande[1], A. Snarskii[2]*

[1]Elvisti, Kiev, Ukraine
[2]Kiev Polytechnical Institute, Kiev, Ukraine

In the task of information retrieval the term relevance is taken to mean formal conformity of a document given by the retrieval system to user's information query. As a rule, the documents found by the retrieval system should be submitted to the user in a certain order. Therefore, a retrieval perceived as a selection of documents formally solving the user's query, should be supplemented with a certain procedure of processing a relevant set. It would be natural to introduce a quantitative measure of document conformity to query, i.e. the relevance measure. Since no single rule exists for the determination of the relevance measure, we shall consider two of them which are the simplest in our opinion. The proposed approach does not suppose any restrictions and can be applied to other relevance measures.

The first relevance measure is determined by frequencies of entry of retrieval terms from the query to the document and is described by the following relation:

$$F(n) = \frac{1}{F_{max}} \sum_{k=1}^{K} M_k, \qquad (1)$$

where $M_k$ is the number of entries of $k$-th term ($k = 1,...,K$) into $n$-th document ($n = 1,...,N$), and $F_{max}$ is determined as follows:

$$F_{max} = \max_{n=1,...,N} F(n). \qquad (2)$$

Note that with one-term query the measure $F(n)$ is equal to normalized in $F_{max}$ number of entries of given term into the document. It is obvious that $F(n) \in (0,1]$.

The second relevance measure includes normalization in the document length $L(n)$:

$$Q(n) = \frac{1}{Q_{max} L(n)} \sum_{k=1}^{K} \ln(M_k + 1), \qquad (3)$$

where $Q_{max}$ is determined as follows:

$$Q_{max} = \max_{n=1,...,N} Q(n), \qquad (4)$$

where $Q(n)$ also varies within $(0,1]$.

The difference in relevance measures in terms of retrieval efficiency in due time was widely covered in the literature [1-3]. It is essential that (1) is a single-parameter measure, and (3) is a double-parameter measure.

Research was made on a set of document bodies including new messages of various size formed by Internet resource monitoring system. As an example, we will consider in this paper an array of 6713 documents published in online media in the first decade of August, 2007, solving the query, presumably on military topics, consisting of one word "military".

Let us first sort out the documents obtained as described above in the order of decreasing relevance measure (abscissa axis), i.e. each document is assigned with an ordinal number $n(Q)$, increasing with decreasing $Q$. Dependence on the number $n(Q)$ is shown in Fig. 1 (smooth curve). As can be seen, a small number of documents have a large relevance measure, and a large number – a small one. Moreover, if we drop the documents with the largest and lowest relevance measures, then dependence of $Q$ on $n(Q)$ in a logarithmic scale to a good accuracy is a linear one (Fig. 2 is a continuous curve), which corresponds to the generalized Zipf law [4, 5] applied to distribution of arbitrary objects in latent variables (properties).

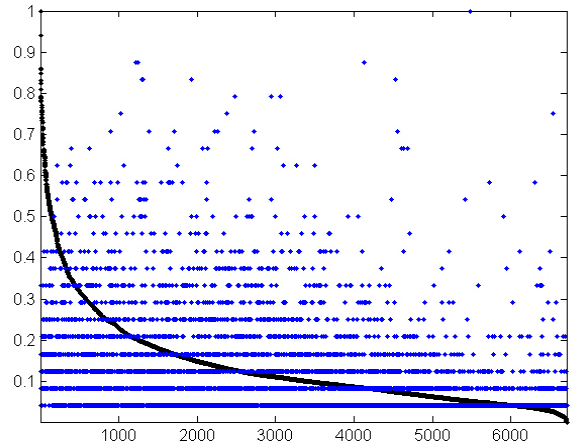

*Fig. 1. The values of relevance measures (axis Y ) of documents based on two criteria. The documents (axis X ) are ranked in values Q .*

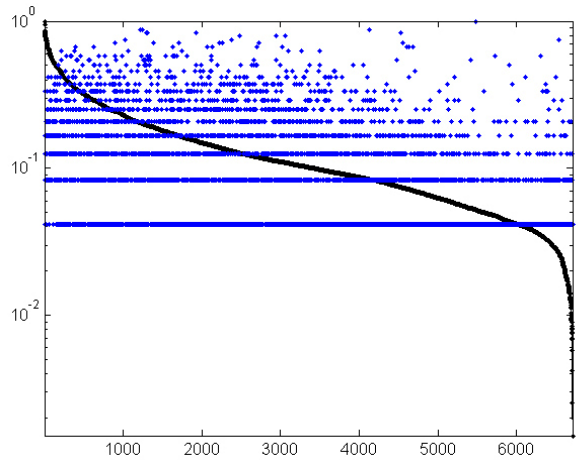

*Fig. 2. The values of relevance measures of documents based on two criteria in a semi-logarithmic scale*

Consider now a dependence of $F$ on $n(Q)$, i.e. leaving "in situ" the documents ranked as $Q$ on the abscissa axis, we will lay on the ordinate not only relevance $Q(n)$, but also relevance $F$. **The thus obtained dependence $F[n(Q)]$ will be referred to as a sequence of mutual relevances.** Behaviour of this dependence given in Fig. 1 and 2, as will be shown below, is characterized by peculiarities typical for a deterministic chaos [6, 7].

For the resulting dependence $F[n(Q)]$ there were calculated factor $D(n)$ by DFA method (Detrended Fluctuation Analysis) and the Hurst index.

The DFA method [8, 9] is a variant of dispersion analysis allowing to study the effects of long-term correlations in the unsteady series with the mean-square error of linear approximation analyzed as a function of approximated portion area.

Within this algorithm of DFA determination the data is first reduced to zero average (subtraction of the average area $\langle F \rangle$ from the time series $F(n)$ ($n=1,...N$) and random walk is constructed:

$$y(k) = \sum_{n=1}^{N}[F(n) - <F>_N]. \qquad (6)$$

Then a series of values $y(k)$, $k=1,...N$ is broken into non-overlapping sections (portions) of length $n$, within each of them the least squares method is used to determine the equation of a line approximating sequence $y(k)$. The resulting approximation $y_n(k)$ ($y_n(k) = ak + b$) is considered as a local trend. In so doing, coefficients $a$ and $b$ are calculated in the following standard way:

$$a = \frac{n\sum ky(n) - (\sum k)(\sum y(k))}{n\sum k^2 - (\sum k)^2};$$

$$b = \frac{(\sum y(k))(\sum k^2) - (\sum k)(\sum ky(k))}{n\sum k^2 - (\sum k)^2}. \qquad (7)$$

Then the mean-square error of a linear approximation is calculated in a wide range of values $n$. It is considered that dependence $D(n)$ is often of power nature $D(n) \sim n^\alpha$, i.e. the presence of a linear portion in a log-log scale $\lg D(\lg n)$ suggests the presence of scaling. With the use of DFA method for different portions of a series of observations of equal length $n$ of sequence under study a linear approximation is constructed for which a mean-square error $D(n)$ is then calculated.

As can be seen from Fig. 3, $D(n)$ has a power dependence on $n$, i.e. in a log-log scale this dependence is close to linear, suggesting the presence of scaling [8, 9].

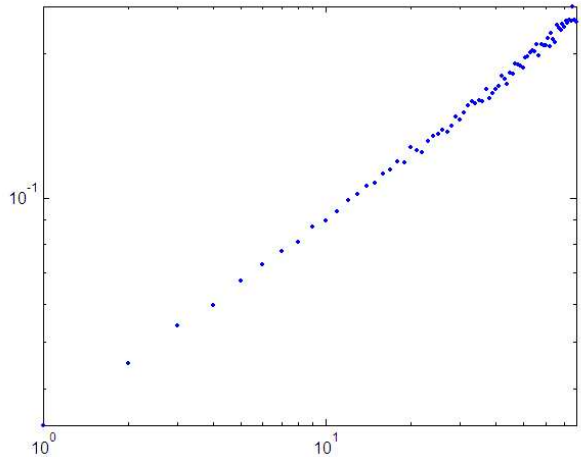

*Fig. 3. D(n) of a series of observations (axis Y) versus the length of approximation portion n (axis X) in a logarithmic scale*

One of the basic characteristics of series characterized by long-term correlations and/or a deterministic chaos is the Hurst index $H$. According to [6] it is found from the relationship:

$$\frac{R}{S} = \left(\frac{N}{2}\right)^H, N \gg 1. \tag{8}$$

Here $S$ is standard deviation:

$$S = \sqrt{\frac{1}{N}\sum_{n=1}^{N}[F(n) - <F>_N]^2}, \tag{9}$$

$$<F>_N = \frac{1}{N}\sum_{n=1}^{N}F(n), \tag{10}$$

and $R$ is the so-called swing:

$$R(N) = \max_{1 \le n \le N} X(n,N) - \min_{1 \le n \le N} X(n,N), \tag{11}$$

where

$$X(n,N) = \frac{1}{N}\sum_{n=1}^{N}[F(n) - <F>_N]. \tag{12}$$

Fig. 4. shows the process of calculation of the Hurst index $H$. With increasing $n$, $H$ index takes on the values $0.75 \div 0.85$. As is known [6], the more the value of $H$ exceeds ½, the more the statistics of series under study differs from the gaussian one, and this series exhibits persistence (the existence of long-term correlations that can be due to the existence of deterministic chaos). If $F(n)$ is taken as a self-affinity function (this issue calls for special investigation), then in conformity with [6] function $F(n)$ has fractal dimension $D$ equal to

$$D = 2 - H \approx 1.25 \div 1.15. \tag{13}$$

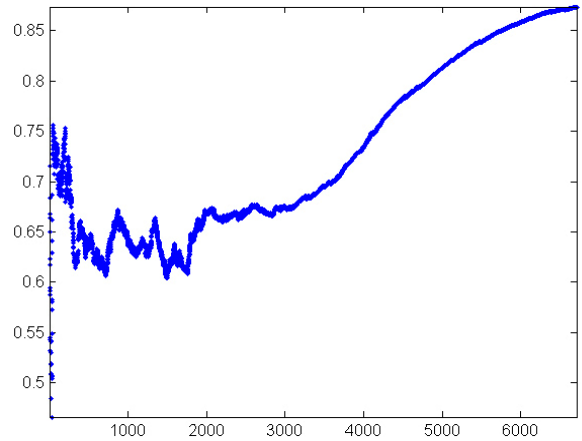

*Fig. 4. The values of the Hurst index (axis Y) versus the size of array under study (axis X)*

Fig. 5 represents the Poincare section of series $F[Q]$ (dependence of $F(n+1)$ on $F(n)$ for $n = 1,...N-1$). Irregular character of filling the square under consideration $(0,1] \times (0,1]$ testifies to the presence of autocorrelations in the series under study.

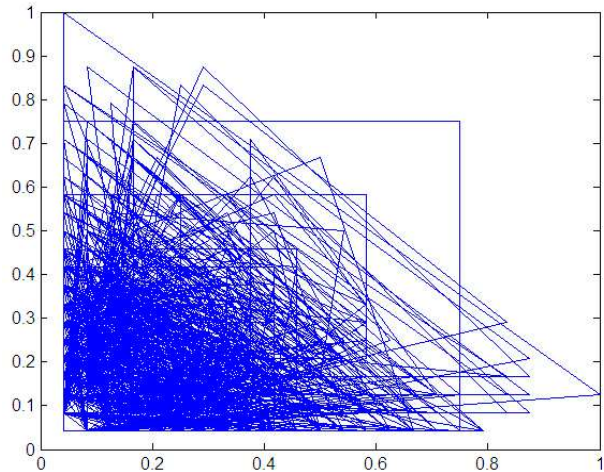

*Fig. 5. Poincare section of the series under study*

The presence of deterministic chaos elements when studying mutual relevance can be also discovered in the construction of a "reverse" series $Q = Q[F]$ (Fig. 6), when the documents are ranked according to relevance measure $F$.

The regularities revealed can be of applied character, for example, for the task of optimization of searching by a set of criteria (for example, by two relevances considered in the paper). From the character of dependence $Q = Q[F]$ it follows that a similar task can be solved by selection of document subsets from the local intervals lying to the right of points belonging to central portion of Fig. 6. The length of these intervals can be determined with regard to restrictions to full sample size. Really, for instance, in the initial portion of the plot large values $Q(n)$ are compensated by smaller values $F(n)$. Investigation of the central portion of data given in Fig. 6 in a sense can play the role of a graphic method of solving the task of optimization of samples obtained by means of multi-parameter criteria.

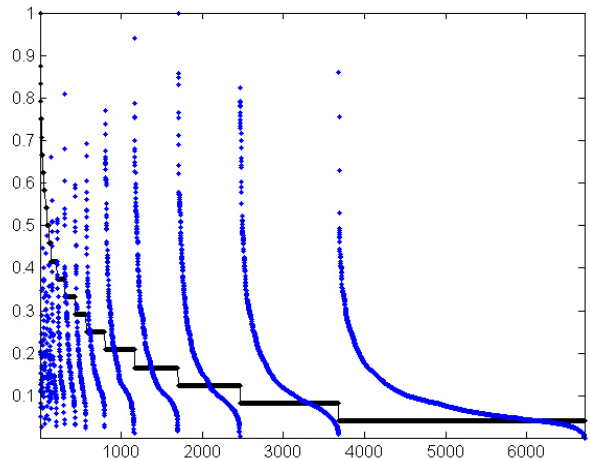

*Fig. 6. The relevance measures of documents based on two criteria.*
*The documents are ranked by Q values*

Regularities reflected in this paper can be also observed in the analysis of other mutual characteristics of information flows, for instance, in the consideration of popularity of subjects of messages in local and global time intervals, or with "unclear" classification of the documents under consideration.

We are grateful to Yandex for the support of our research under Grant "Internet Mathematics

2007".